\documentclass[runningheads]{llncs}
\usepackage[T1]{fontenc}
\usepackage{times}    
\usepackage{amsmath}  
\usepackage{amssymb}  
\usepackage{amstext}
\usepackage{graphicx}
\usepackage{multicol}
\usepackage[bookmarks=true]{hyperref}
\usepackage[usenames]{color}
\usepackage[x11names, table]{xcolor}
\usepackage{xspace}
\usepackage{balance}
\usepackage[lofdepth,lotdepth]{subfig}
\usepackage{pifont}
\usepackage{derivative}
\usepackage{algorithm}
\usepackage{algpseudocode}
\usepackage[multi-part-units=single]{siunitx}
\usepackage{enumitem}

\usepackage{multirow}
\usepackage{multicol}
\usepackage{makecell}
\usepackage{tabularx}
\usepackage{array}

\usepackage{graphicx}
\usepackage{subcaption}
\usepackage{pgfplots}
\pgfplotsset{compat=1.18}
\usepackage{pgfplots}
\usepgfplotslibrary{fillbetween}
\usepackage{keyval}
\usepackage{ifthen}
\usepackage[dvipsnames]{xcolor}
\usepackage[export]{adjustbox}
\usepackage{tikzscale}
\usepackage{svg}
\usepackage{svg-extract}

\newcommand{\tx}{\text}
\newcommand{\RRR}{\mathbb{R}}

\newcommand{\TT}{\mathcal{T}}
\newcommand{\NN}{\mathcal{N}}
\renewcommand{\AA}{\mathcal{A}}
\renewcommand{\t}{\theta}
\newcommand{\Exp}[2][]{\mathbb{E}_{#1}\!\left\{#2\right\}}
\usepackage{mathtools}
\DeclarePairedDelimiter{\norm}{\Vert}{\Vert}

\usepackage{comment}
\newif\ifdisclosure

\disclosuretrue

\ifdisclosure
\includecomment{disclosure}
\excludecomment{anonymous}
\else
\includecomment{anonymous}
\excludecomment{disclosure}
\fi

\def\BibTeX{{\rm B\kern-.05em{\sc i\kern-.025em b}\kern-.08em
    T\kern-.1667em\lower.7ex\hbox{E}\kern-.125emX}}
\begin{document}

\title{Masked Registration and Autoencoding of CT Images for Predictive Tibia Reconstruction}
\titlerunning{Predictive Tibia Reconstruction}


\begin{disclosure}
    \author{
        Hongyou Zhou\index{Zhou, Hongyou}\inst{1} \and
        Cederic Aßmann\index{Aßmann, Cederic}\inst{1} \and
        Alaa Bejaoui\index{Bejaoui, Alaa}\inst{2} \and
        Heiko Tzschätzsch\index{Tzschätzsch, Heiko}\inst{2} \and
        Mark Heyland\index{Heyland, Mark}\inst{4} \and
        Julian Zierke\index{Zierke, Julian}\inst{4} \and
        Niklas Tuttle\index{Tuttle, Niklas}\inst{3} \and
        Sebastian Hölzl\index{Hölzl, Sebastian}\inst{3} \and
        Timo Auer\index{Auer, Timo}\inst{3} \and
        David A. Back\index{Back, Alexander David}\inst{3} \and
        Marc Toussaint\index{Toussaint, Marc}\inst{1}
    }

    \authorrunning{Zhou and Toussaint et al.}
    \institute{
    Technical University Berlin, Learning and Intelligent Systems, Berlin, Germany 
    \and
    Charité – Universitätsmedizin Berlin, corporate member of Freie Universität Berlin and Humboldt-Universität zu Berlin, Institute of Medical Informatics, Berlin, Germany 
    \and 
    Charité – Universitätsmedizin Berlin, corporate member of Freie Universität Berlin and Humboldt-Universität zu Berlin, Center for Musculoskeletal Surgery, Berlin, Germany
    \and
    Julius Wolff Institute, Berlin Institute of Health at Charité – Universitätsmedizin Berlin, Germany}
\end{disclosure}

\begin{anonymous}
    \author{Anonymized Authors}
    \authorrunning{Anonymized Author et al.}
    \institute{Anonymized Affiliations \\
        \email{email@anonymized.com}}
\end{anonymous}

\maketitle


\begin{abstract}

Surgical planning for complex tibial fractures can be challenging for surgeons, as the 3D structure of the later desirable bone alignment may be difficult to imagine.
To assist in such planning, we address the challenge of predicting a patient-specific reconstruction target from a CT of the fractured tibia.
Our approach combines neural registration and autoencoder models.
Specifically, we first train a modified spatial transformer network (STN) to register a raw CT to a standardized coordinate system of a jointly trained tibia prototype.
Subsequently, various autoencoder (AE) architectures are trained to model healthy tibial variations.
Both the STN and AE models are further designed to be robust to masked input, allowing us to apply them to fractured CTs and decode to a prediction of the patient-specific healthy bone in standard coordinates.
Our contributions include: i) a 3D-adapted STN for global spatial registration, ii) a comparative analysis of AEs for bone CT modeling, and iii) the extension of both to handle masked inputs for predictive generation of healthy bone structures. Project page: \href{https://github.com/HongyouZhou/repair}{https://github.com/HongyouZhou/repair}

\end{abstract}

\section{Introduction}
\begin{figure}[t]
\centering
\includegraphics[width=\linewidth]{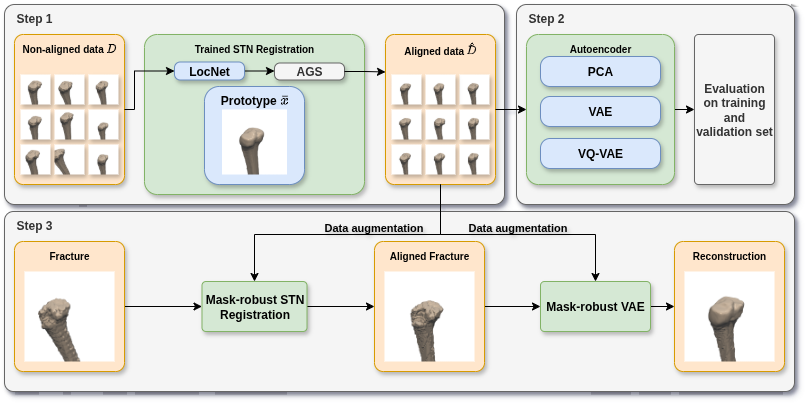}
\caption{\textbf{Overview of the proposed methods}. (See details in Sec.~\ref{sec:methods}.) Top left: Training data $D$ of non-aligned CT images. Step 1 trains an STN and prototype jointly for consistent registration. Step 2 trains alternative AEs to provide latent representations of individual tibia. Step 3 retrains STN and AEs to become masked-robust and enable predicting a tibia reconstruction.}
\label{fig:overview}
\end{figure}

Fractures are a relevant burden for all societies around the world, with a notable increase in numbers internationally within the last decades~\cite{wu2021global}. 
Surgical interventions are often indicated, especially when larger bones and joints are involved~\cite{jones2020principles}. 
In the joint area in particular, exact restoration of the initial anatomical situation and maintenance of alignment and length by surgical fixation are essential to prevent early osteoarthritis~\cite{schenker2014pathogenesis}. 
For this reason, planning is of decisive importance with regard to adequate fragment repositioning, possible bony supplementation, and final osteosynthesis based on the prediction of the ideal result~\cite{scolozzi2012maxillofacial,knopf2001adaptive}.
However, planning such operations based solely on computer tomography (CT) imaging poses significant challenges. The difficulty in reconstructing the desired bone alignment, especially in multi-fragmentary situations, often lead to post-traumatic arthrosis, especially if the joint surface is not reconstructed accurately~\cite{keil2023intraoperative}.

Predicting the entire healthy structure from partial input (e.g.\ fracture-masked) can be considered an instance of (3D) image completion~\cite{liu2018imageinpaintingirregularholes}. 
Previous approaches~\cite{liu2018imageinpaintingirregularholes} in computer vision include using U-Net-based~\cite{ronneberger2015u} architectures to complete the partially masked images. 
More recently, a framework called Resolution-Robust Large Mask Inpainting (LaMa)~\cite{suvorov2022resolution} has been introduced with fast Fourier convolutions (FFC)~\cite{chi2020fast} for image completion.
In general, Masked Autoencoders (MAE)~\cite{kingma2013auto,he2022masked} are a
strong approach to learn latent representations that are robust to
partially masked information. 
The encoder operates only on the visible patches and generates latent codes, which are then
processed by the decoder along with mask tokens to reconstruct the
target image.
In our approach, we study various AE architectures for 3D image tensors, trained to directly reconstruct masked CT images.

However, naively applying autoencoders (AEs) on raw tibia CTs performs poorly, as the latter are commonly not aligned to standardized coordinates.
Structure registration -- that is, identifying the spatial transform of some structure to standard coordinates -- has extensively been studied in medical imaging~\cite{miao2016cnn,miao2018dilated} and many other areas~\cite{peng2019pvnet,deng2020self}.  
To avoid issues with local optima of traditional methods like Iterative Closest Point (ICP)~\cite{besl1992method}, modern approaches train neural nets to directly predict the spatial transform from the image input: For 2D image data, Spatial Transformer Network (STN) architectures~\cite{lee2019image,jaderberg2016spatialtransformernetworks} show impressive results for deformable object registration. 
VoxelMorph~\cite{balakrishnan2019voxelmorph} is a U-Net-based model to align a moving against a fixed 3D volumetric data from CT scans,
and C2FViT~\cite{mok2022affine} applied attention mechanisms, using Vision Transformers (ViT)~\cite{dosovitskiy2020image}.
Further deep learning-based registration methods include~\cite{choy2020deep,zhou2016fast,kim2022diffusemorph}. 
However, the above studies only address registration against a known target, whereas our focus lies on inferring a consistent prototype from an entire dataset as detailed below.

In our work, we propose a modified STN that is trained to predict a registration of a raw tibia CT to shared coordinates, such that they align best with a co-optimized \emph{prototype} tibia CT. 
We then consider various AE to model the distribution of tibia varieties, essentially modeling the variance around the prototype. 
Both, the STN-registration as well as the AE are then trained to be mask-invariant, to be applied to a fracture-masked tibia and to decode it to the predicted reconstruction.


Our novel contributions include: i) the modified STN to predict the global spatial registration of a 3D image tensor, 
ii) a comparison of alternative AE architectures to model bone CTs, 
and iii) an extension of both of the above to masked CT images, enabling the predictive decoding to a healthy bone in standard coordinates.

\section{Methods}
\label{sec:methods}
In this section, we now present our method for end-to-end reconstruction.
Figure~\ref{fig:overview} outlines our method, which builds on training data $\mathcal{D}=\{(x_i,m_i)\}_{i=1}^N$ of $N$ CT scans of healthy tibiae (Fig.~\ref{fig:overview}, top left). 
Each CT scan $x_i \in \mathbb{R}^{W \times H \times S}$ is a single-channel tensor.
We further assume to have a binary mask $m_i\in \{0,1\}^{W \times H \times D}$ indicating the tibia bone voxels while masking out neighboring bones like the femur, fibula, patella.
At query time, we are given a fractured tibia CT (Fig.~\ref{fig:overview}, bottom left) with fractures coarsely masked out and aim to predict its reconstruction and affine transform to standard coordinates. To this end, we train models in the following steps:
\begin{enumerate}[label=\textsc{Step \arabic*:}, wide]
    \item We jointly train a tibia prototype tensor $\bar x$ and a Spatial Transformer Network \cite{jaderberg2016spatialtransformernetworks} that predicts an affine transformation to match the input tibia CTs best to the prototype. Output of this step is the tibia prototype $\bar x$ and a consistently aligned data set $\hat{\mathcal{D}}=\{(\hat x_i,\hat m_i)\}_{i=1}^N$.
    \item We train autoencoder models on the aligned dataset to capture the variance of healthy tibia around the prototype in their latent code (Fig.~\ref{fig:overview}, top right).
    \item We then train both, the STN and autoencoders to become robust to masking, so that they can also be applied on a fracture-masked tibia CT, mapping it first to standard coordinates, then a latent code, and then decoding to its reconstruction (Fig.~\ref{fig:overview}, bottom).
\end{enumerate}

\subsection{Joint Prototype and Registration Net Optimization} \label{sec:proto}

An essential part of a \emph{Spatial Transformer Net (STN)}~\cite{jaderberg2016spatialtransformernetworks} is a transformation $\hat x = \AA(x, \theta)$ which performs a \emph{differentiable} affine transformation by grid re-sampling an image or 3D tensor $x$ for given transform parameters $\t$.\footnote{If $x$ was not discretized and $\TT_\t$ is the affine transformation, we had $\hat x(p) = x(\TT_\t^{-1}(p))$ for $p\in\RRR^3$.
An AGS $\hat x = \AA(x, \theta)$ approximates this operation for a discretized tensor $x$ using tri-linear interpolation and providing differentiation w.r.t.\ both $x$ and $\t$.}
We call $\AA$ \emph{Affine Grid Sampler (AGS)}. 
This differentiability can be leveraged to directly optimize for individual affine parameters $\t_i$ for each data CT $x_i$, or a \emph{Localization Network (LocNet)}~\cite{jaderberg2016spatialtransformernetworks} can be trained to predict $\t$ as a function of $x$. We evaluate both approaches. 
Using a LocNet can be less prone to local optima and will allow us to directly generalize to potentially masked query CTs. However, we consider directly using AGS for registration as an evaluation baseline, which we explain first for simplicity:

\paragraph{Direct AGS Registration:} We solve
\begin{align}\label{eq_reg_p}
    \min_{\bar x, \theta_{1:N}}~ \textstyle\sum_{i=1}^{N} \ell(\bar x, \mathcal{A}(x_i \cdot m_i, \theta_i)) ~,
\end{align}
where $\bar x$ becomes the tibia prototype tensor, $\t_i$ are individual transforms for each data CT $x_i$, $x\cdot m$ is the elem-wise multiplication, and $\ell$ is a loss function comparing two CT tensors (we use D-SSIM). 

\paragraph{STN Registration:} We solve
\begin{align}\label{eq_reg_p_locnet}
    \min_{\bar x, \phi}~ \textstyle\sum_{i=1}^{N} \ell(\bar x, \mathcal{A}(x_i \cdot m_i, \tx{LocNet}_{\phi} (x_i \cdot m_i))) ~,
\end{align}
and use a modified $\tx{LocNet}_{\phi}$, which is based on 3D convolutions, see Fig.~\ref{fig:overview} that maps the masked data CT to affine transform parameters; $\t$ are the network parameters of $\tx{LocNet}_{\phi}$. 

\subsection{Autoencoders and Healthy Tibia Latent Codes}

A \emph{variational autoencoder} (VAE) trains an encoder $q_\phi(z|x)$ and decoder $p_\theta(x'|z)$ to minimize
\begin{align}\label{eq:vac}
  \min_{\phi,\theta}~ \Exp[x\sim \mathcal{D}, z \sim q_\phi(z|\hat x)]{ \ell(x', \hat x) - \beta D_\text{KL}\left(q_\phi(z| \hat x) \,\|\, p(z) \right) }~,
\end{align}
where $\ell(x',\hat x)$ is the reconstruction error between the aligned data $\hat x$ and decoded $x'$, and $p(z) = \NN(0, \mathrm{I})$ is a latent code prior. Training an autoencoder implies training a latent code that represents the variance of $\hat x$, such that we can encode $\hat x\mapsto z$ and decode $z\mapsto x'$. 
Note that this VAE is trained on unmasked, aligned CTs, while a mask-robust version is later retrained with masked loss \ref{eq:vae_masked} to support fracture completion.
In step 2 of our pipeline, we train autoencoders on our aligned and masked data CTs $\{\hat x_i \circ \hat m_i\}_{i=1}^N$. 
We compare three alternative approaches in this work:

\paragraph{VAE:} We train a standard variational autoencoder minimizing \eqref{eq:vac}, where the latent code maintains spatial consistency, i.e. $z \in \RRR^{E_{\tx{VAE}} \times \frac{WHS}{L^3}}$, where $E_{\tx{VAE}}$ is the predefined dimension of the latent space and $L$ is the total level of downsampling. The specific network used in this work is a U-Net-based architecture~\cite{ronneberger2015u,he2016deep} 
%

\paragraph{Vector Quantized VAE:}
A Vector Quantized VAE (VQ-VAE) provides an alternative latent structure to capture data variance, where the latent variable $z$ is categorical, i.e., discrete.
%
We define a latent embedding space $z \in \RRR^{E_{\tx{VQVAE}} \times \frac{WHD}{L^3}}$ where $E_{\tx{VQVAE}}$ is the dimensionality of each latent embedding vector $z_i$, and a learnable codebook $C \in \RRR^{K \times E_{\tx{VQVAE}}}$ which has $K$ vectors.
The model takes a aligned input CT image $\hat x$, through an encoder, resulting in the output $z_e(\hat x)$. The discrete latent variables $z_i$ are then calculated by a nearest neighbor lookup using the vectors in $C$. 

\paragraph{PCA:} Making the encoder and decoder deterministic and linear with limited rank $p$, and choosing the loss function as square error and $\lambda=0$, we retrieve PCA with $p$ principal components. We include this approach as a classical baseline to model the tibia CT variance around the prototype. 
%
%
%
We will evaluate these three autoencoder approaches, with the PCA being our baseline, for their performance in reconstructing healthy tibia CTs.

\subsection{Extension to Fracture-Masked CTs via Data Augmentation} \label{sec:data_aug}
At application time, we have a query CT $y$ of a fractured tibia together with a mask $m$ that (coarsely) masks out the fractured parts. Our system aims to predict the reconstruction as well as affine parameters $\t$ to transform to standard coordinates.

Applying the previously trained LocNet directly on the facture-masked CT performs mediocre, as it has never been trained on heavily masked CTs (see Table~\ref{tab:qual_comp_val_set} for masking examples). Therefore, in step 3 we train a second, \emph{mask-robust} LocNet using data augmentation. To this end, we start with our aligned training data $\hat{\mathcal{D}} = \{(\hat x_i, \hat m_i)\}_{i=1}^N$ which is the output of step 1. We now consider these $\hat x_i$ as ground truth data and generate a large data set of randomly masked and transformed instances of these $\hat x_i$ to train the second LocNet to reversely predict these transformations.

More specifically, we apply random masks $m \sim \mathrm{Bernoulli}(m)$ sampled from a distribution over blocks of voxels with random size in $[10,20]$, and sample random affine transforms $\t\sim p(\t)$ uniformly with Euler angles $\le 15^\circ$, translations $\le 2.5cm$ and scalar scaling $[0.9,1.1]$. A data CT $\hat x$ is therefore transformed to $\tilde x = \TT_{\t}^{-1}(\hat x \cdot m)$. We train the second LocNet to predict this transformation by minimizing
\begin{align}\label{eq_frac_reg}
    \min_{\phi}~ \Exp[\hat x \sim \hat D,p(m), p(\t)]{ \ell(\hat x \cdot m, \mathcal{A}(\tilde x, \text{LocNet}_{\phi} (\tilde x))) + \alpha  \norm{\t - \text{LocNet}_\phi(\tilde x)}^2 } ~,
\end{align}
where the second term of the loss function is a standard square error regression loss to the true affine transform. During the experiments, we trained all our models from scratch.

Finally, we also train our autoencoders alternatives from scratch to become mask-robust, using the masked reconstruction loss as follow:
\begin{align}\label{eq:vae_masked}
  \min_{\phi,\theta}~ \Exp[x\sim D, z \sim q_\phi(z|\hat x)]{ \ell(x' \cdot \neg m,\hat x \cdot \neg m) - \beta D_\text{KL}\left(q_\phi(z| \hat x \cdot \neg m) \,\|\, p_\theta(z) \right) }~,
\end{align}
where the first term differs from~\ref{eq:vac} with only the non-masked area been used for calculating the loss.

\section{Experiments}


\newcommand{\maxf}[1]{{\cellcolor[gray]{0.8}} #1}
\begin{table}[t]
\renewrobustcmd{\bfseries}{\fontseries{b}\selectfont}
\centering
\caption{Reconstruction performances (mean with $95\%$ confidence interval) for the investigated registration and autoencoding methods on a left-out validation dataset.
Autoencoding method $P$ means predicting the constant prototype (directly relates to losses \eqref{eq_reg_p} and \eqref{eq_reg_p_locnet}), respectively; while the performances for other autoencoding methods relate to \eqref{eq:vac}. The bottom block evaluates mask-robust methods on randomly masked samples of the validation dataset.}
\label{tab:quat_comp_val_set}
\sisetup{
separate-uncertainty=true,
round-mode=uncertainty, 
group-separator=\text{},
round-precision=1,
multi-part-units=single,
bracket-numbers = false,
print-zero-integer = true,
detect-all=true,
detect-inline-weight=math
}
\resizebox{\linewidth}{!}{%
\begin{tabular}{|c|c| 
S[scientific-notation=fixed,fixed-exponent=-2, table-format=1.3(1.3)] 
S[scientific-notation=fixed,fixed-exponent=-2, table-format=1.3(1.3)] 
S[scientific-notation=fixed,fixed-exponent=-2, table-format=1.3(1.3)] 
S[scientific-notation=fixed,fixed-exponent=-2, table-format=1.3(1.3)] 
|
S[table-format=2.2(1.2)] 
S[scientific-notation=fixed,fixed-exponent=-2, table-format=1.3(1.3)] 
S[table-format=1.3(1.3)]
|}
\hline
 
Registration & Autoencoding & \multicolumn{4}{c|}{D-SSIM $\times 10^{-2}$} & && \\
Method &
Method &
{$\tx{D-SSIM}_{15} \downarrow$} &
{$\tx{D-SSIM}_{11}\downarrow$} &
{$\tx{D-SSIM}_{7}\downarrow$} &
{$\tx{D-SSIM}_3\downarrow$} &
{$\tx{PSNR}\uparrow$} &
{$\tx{RMSE}\downarrow \times 10^{-2}$} &
{$\tx{RMSD}\downarrow$}\\
\hline
{\multirow{5}{*}{\makecell{w/o \\ Registration}}}  
& P   & N/A & N/A & N/A & N/A & N/A & N/A & N/A\\ 
& $\tx{PCA}_{32}$ & 
0.01664875613318549(0.0007912259296683309) & 
0.016647324517921166(0.0007898485529793968) & 
0.016689325924272896(0.0007566921572770006) & 
0.017105474516197476(0.0005964196563127651) & 
31.991376311690715(0.35830097039510245) & 
0.025455816855861076(0.003398967530621464) & 
1.5949316473803754(0.13023528405505114)\\ 
& $\tx{PCA}_{16}$ & 
0.016985931882151844(0.0008365385186977965) & 
0.01698143228336617(0.0008353968375617851) & 
0.016947547594706215(0.0008061797522638038) & 
0.016957825532665963(0.0006645116966271725) & 
31.463969972398544(0.3563413692818919) & 
0.02704199617383656(0.0037492360968660807) & 
1.5884868517511581(0.13537030195568564)\\ 
& VAE & 
0.007190997557093699(0.00034211831242498555) &
0.007211656723585393(0.00033997894152110886) & 
0.007289892735166683(0.00034175782447836816) & 
0.00772983513565527(0.00036338943975606434) & 
33.065775553385414(0.36990255277776757) & 
0.022648220881819725(0.004717212861609128) & 
0.7399625664813341(0.03247699874245162)\\ 
& VQ-VAE & 	
\bfseries 0.0014544372679665685(0.00009338633558732737) & 	
\bfseries 0.001456961061598526(0.00009348113906498797) & 	
\bfseries 0.0015430379038055737(0.00009545097049256508) & 	
\bfseries 0.0026010833163228296(0.00011470780671469184) & 	
\bfseries 43.76724667019314(0.6494134946130806) & 	
\bfseries 0.006848246169586976(0.001110340568190161) & 	
\bfseries 0.7441745011632422(0.021380441143050044)\\   
\hline
{\multirow{5}{*}{\makecell{Direct AGS~\ref{eq_reg_p}\\Registration}}}
& P   & 
0.01135346572846174(0.0004735046701158597) & 
0.01131404777643857(0.0004713639387237148) & 
0.01052123977354279(0.00043263599315175393) & 
0.008160593884962574(0.0003079250545932524) & 
28.29331814801251(0.2743336983180342) & 
0.03900840340389146(0.00719745167912021) & 
2.0235141930611795(0.8703873268857633)\\ 
& $\tx{PCA}_{32}$ & 
0.01633695706173225(0.0007880602611303825) & 
0.016336032085948513(0.0007866982120127982) & 
0.01638820546644705(0.0007533548980669755) & 
0.01683376674298886(0.0005918307370763068) & 
32.34716401276764(0.3698492412887344) & 
0.02445481193286401(0.0031539549231793837) & 
1.56617499336258(0.12809033482952706)\\ 
& $\tx{PCA}_{16}$ & 
0.016760126860053452(0.0008277762120175264) & 
0.01675623600129728(0.0008266375204466819) & 
0.016732495692041185(0.0007975822128124037) & 
0.01676245640825342(0.0006574767190777763) & 
31.784438945628978(0.3666125278204015) & 
0.026081271952501043(0.003493921688341166) & 
1.5687894794425024(0.1336431934810827)\\ 

& VAE & 
0.011574677502115566(0.00064369278643893) & 
0.011579266956282986(0.0006438202452334441) & 
0.011658345038692156(0.0006403313171045792) & 
0.012489944903386964(0.0006199870477620393) & 
40.15658442179362(0.327861765645936) & 
0.010012523064182864(0.0006411456004977963) & 
0.568795047150313(0.008133955507814296)\\   

& VQ-VAE & 	
\bfseries 0.001140505670466357(0.00012333135322701802) & 	
\bfseries 0.0011414256813522014(0.00012349329011092548) & 	
\bfseries 0.0011791245002920432(0.00012660437818623563) & 	
\bfseries 0.0016894926084205506(0.00015145878857139635) & 	
\bfseries 45.76874160766601(0.25035728384401146) & 	
\bfseries 0.005202674048228397(0.00016221666344963274) & 	
\bfseries 0.38406632309453703(0.0074613715065239864)\\  

\hline
{\multirow{5}{*}{\makecell{STN~\ref{eq_reg_p_locnet}\\Registration}}}
& P   & 
0.0036250502043576145(0.000496090911094132) & 
0.0036179037350747316(0.0004939899420505251) & 
0.0034819560004743155(0.0004491611256700097) & 
0.003033307975985937(0.00029935578978056637) & 
35.66410629837601(0.5067043353210079) & 
0.0175191990479275(0.0022159121679690944) & 
0.7719681750030623(0.30453800823402055) \\ 

& $\tx{PCA}_{32}$ & 
0.0026021682553821128(0.00047365022053762093) & 
0.0025992735668464947(0.00047247123175165893) & 
0.002580253614319696(0.00044125209863467993) & 
0.0026331775718265098(0.0003019113333110822) & 
39.49588235219319(0.5596755041440498) & 
0.010959273963062847(0.0014636276014536346) & 
0.6938937416008498(0.11294655714607955)\\ 
& $\tx{PCA}_{16}$ & 
0.0027392484523631896(0.00047062812825493435) & 
0.002735732330216301(0.0004694258352554797) & 
0.0027026815546883475(0.0004379703645521335) & 
0.0026873030044414366(0.0003005437716524351) & 
38.87070980778448(0.5340410003347925) & 
0.01173761423194298(0.0015473381390749565) & 
0.6938267032995306(0.11121547523164899)\\ 

& VAE & 
0.001181295004155902(0.00017046009269404807) & 
0.00118420807768901(0.0001702841030314989) & 
0.00120908794634872(0.00016498369753885784) & 
0.001579695243802335(0.00013174335897814777) & 
42.26615227593315(0.3640199523731436) & 
0.007890183271633254(0.007890183271633254) & 
0.5252590217266043(0.027200714897046886)\\   

& VQ-VAE & 	
\bfseries 0.0006209678378783994(0.00011220383381063144) & 	
\bfseries 0.0006221207408493178(0.00011227553813101833) & 	
\bfseries 0.0006628980433257917(0.00011436952786564886) & 	
\bfseries 0.0010917667273638977(0.00012666321471132275) & 	
\bfseries 51.16865242852105(0.22523856559723482) & 	
\bfseries 0.0027897320946471556(0.00005859414559308857) & 	
\bfseries 0.39472172035524766(0.00516740208326045)\\   
\hline
\hline
{\multirow{5}{*}{\makecell{Mask-robust~\ref{eq_frac_reg}\\STN\\Registration}}}
& P   & N/A & N/A & N/A & N/A & N/A & N/A & N/A\\ 
& $\tx{PCA}_{32}$ & 
0.004249622424443559(0.00019954539715890987) & 
0.0042396220895979135(0.00019889973516680614) &
0.004056279857953387(0.00018465944814718825) &
0.00353793634308709(0.0001288944517001471) &
36.25393705014828(0.4765798399722228) &
0.015713219599867306(0.0017993836352386435) &
0.6710157162299232(0.03207359413791263)\\ 
& $\tx{PCA}_{16}$ & 
0.004404680596457585(0.0002205984671801287) & 
0.004394674190768489(0.00021993522566233464) & 
0.00420772808569449(0.00020508026525064215) & 
0.0036624493422331614(0.0001480871166792448) & 
36.242796933209426(0.466674787487145) & 
0.01572106887275973(0.0017465669539955295) & 
0.6724579791638271(0.03126404095235117)\\ 

& VAE & 
\bfseries 0.0014298437163233757(0.00008090300722387914) & 
\bfseries 0.0014294855645857751(0.00008077349112719437) & 
\bfseries 0.0014381750952452421(0.00007713142205422591) & 
\bfseries 0.0015914186369627714(0.000059420488192200475) & 
43.25596046447754(0.34627400462768554) &
0.007025767117738724(0.00043580540846641395) & 
\bfseries 0.5227560698986053(0.010329485039599929) \\

& VQ-VAE & 	
0.003018034958384103(0.0003290086925340603) &
0.003019261277384228(0.00032908133071834475) & 
0.0030634503345936537(0.0003288607830890938) & 
0.0035465702983654207(0.0003329653247226302) & 
\bfseries 44.03749889797635(0.7235716493297008) & 
\bfseries 0.006809761385536856(0.0004895386424482849) & 
0.6035331036411872(0.043848543521306634)\\   
\hline
\end{tabular}
}

\end{table}


\begin{table}[t]
  \caption{Qualitative comparison of reconstructed tibia for investigated registration and autoencoding methods. 3D models were the isosurfaces generated using a Marching Cubes methods.}
    \label{tab:qual_comp_val_set}
    \centering

\begin{tabular}{|c|cc|ccccc|}
\hline
\multicolumn{3}{|c|}{Registration Method} &
\multicolumn{5}{c|}{Autoencoder Method}
\\
\hline
& 
Input&
Prototype &
$\tx{PCA}_{32}$ &
$\tx{PCA}_{16}$ &
VAE &
VQ-VAE &
Ground Truth \\
\hline
&&&&&&&\\[-2.5ex]
\makecell{w/o\\Registration} &
\includegraphics[width=0.10\linewidth,valign=c]{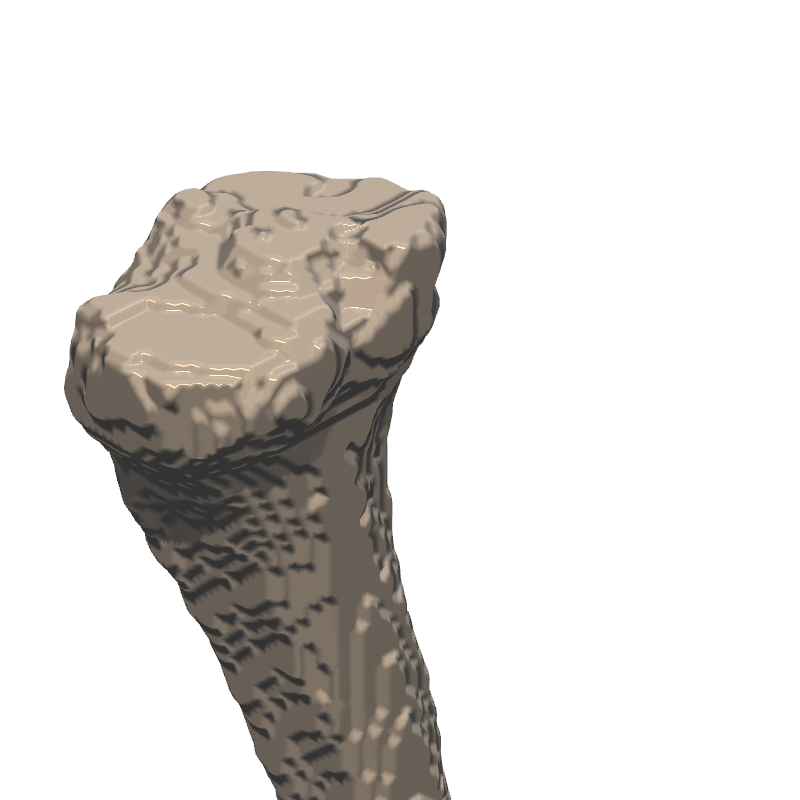} &
N/A &
\includegraphics[width=0.10\linewidth,valign=c]{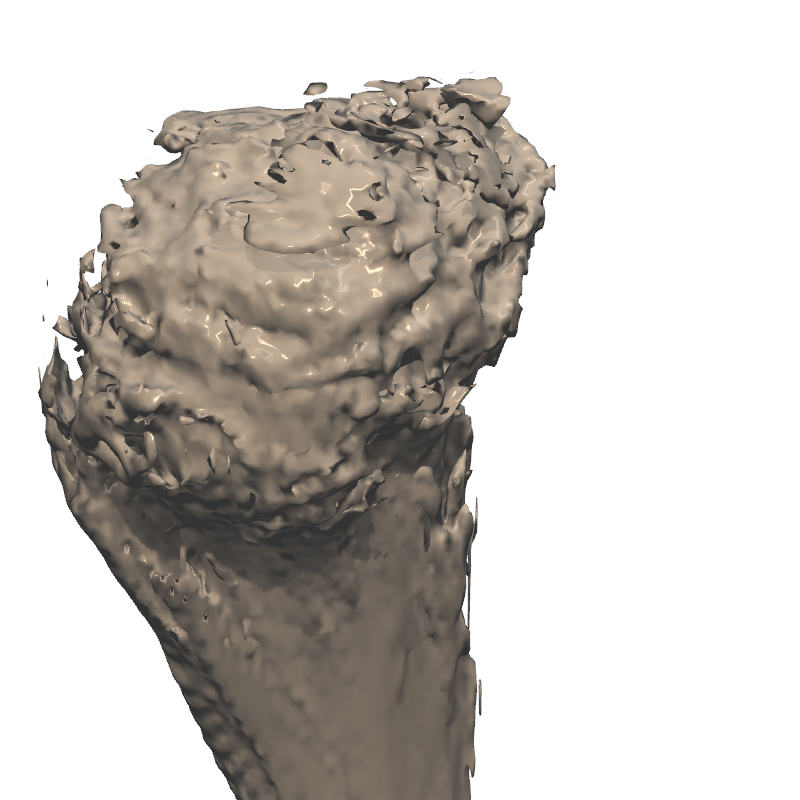} &
\includegraphics[width=0.10\linewidth,valign=c]{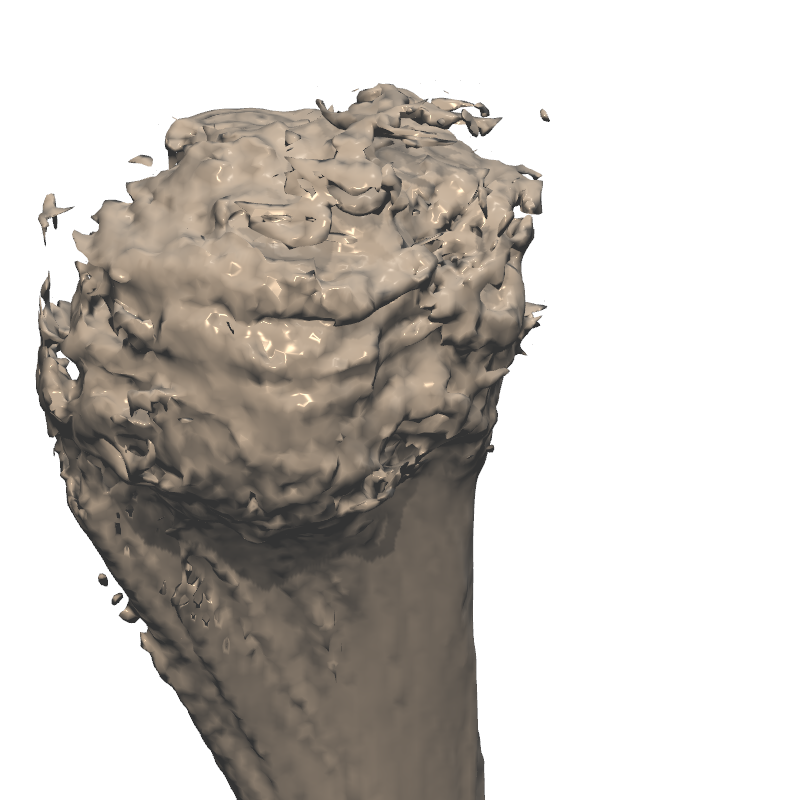} &
\includegraphics[width=0.10\linewidth,valign=c]{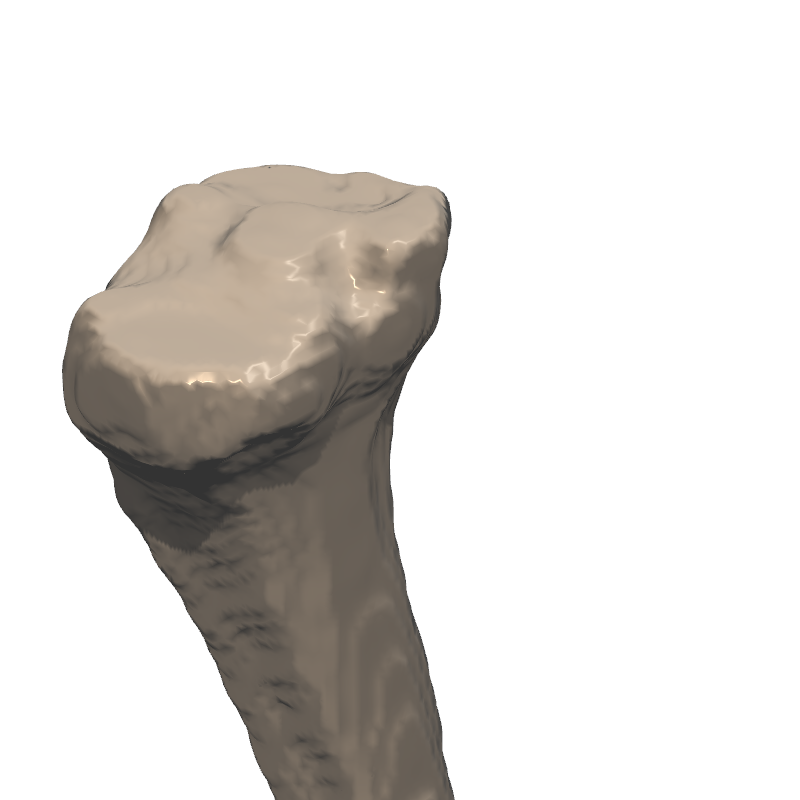} &
\includegraphics[width=0.10\linewidth,valign=c]{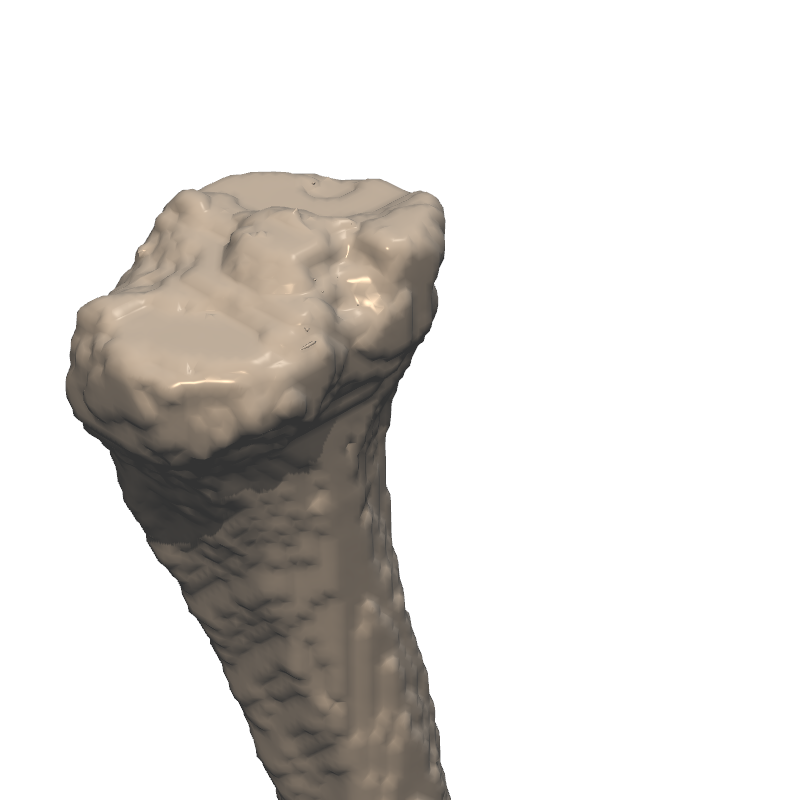} &  
\includegraphics[width=0.10\linewidth,valign=c]{imgs/tibia_no_reg_gt.png}  
\\
\makecell{Direct AGS~\ref{eq_reg_p}\\Registration} & 
\includegraphics[width=0.10\linewidth,valign=c]{imgs/tibia_no_reg_gt.png} &
\includegraphics[width=0.10\linewidth,valign=c]{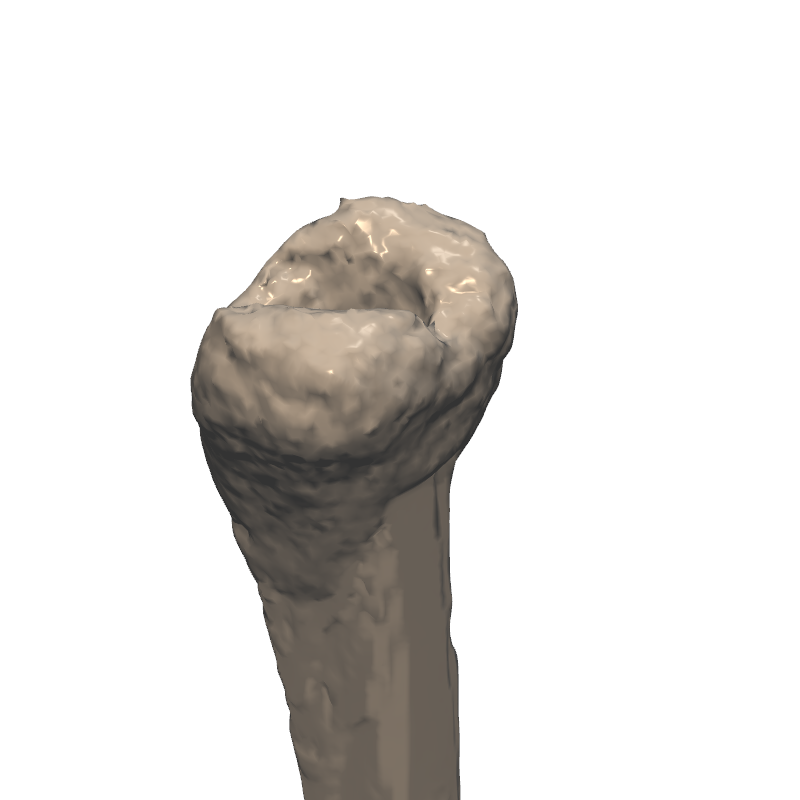} &
\includegraphics[width=0.10\linewidth,valign=c]{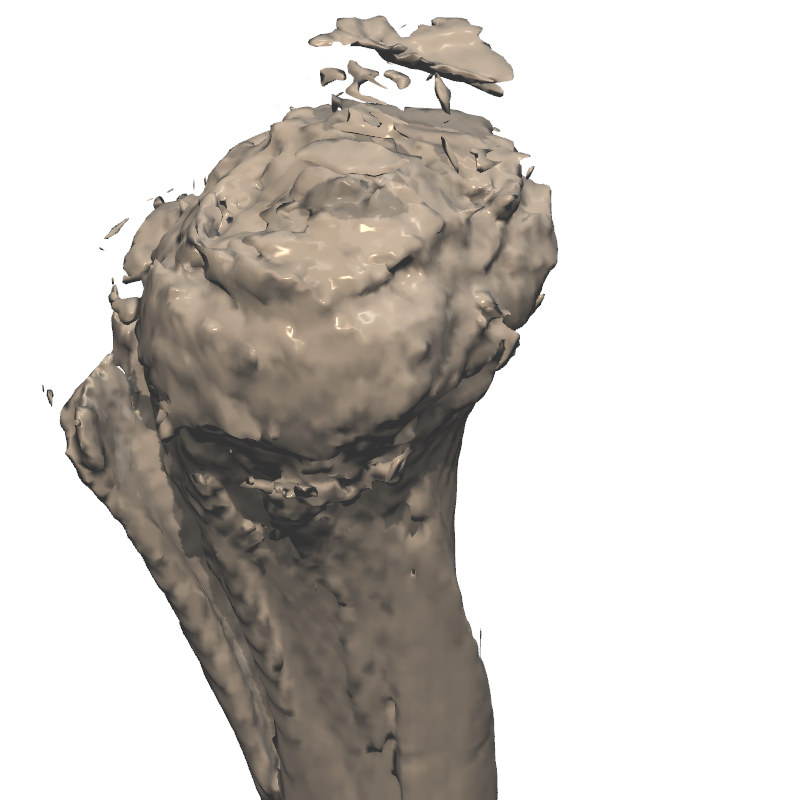} &
\includegraphics[width=0.10\linewidth,valign=c]{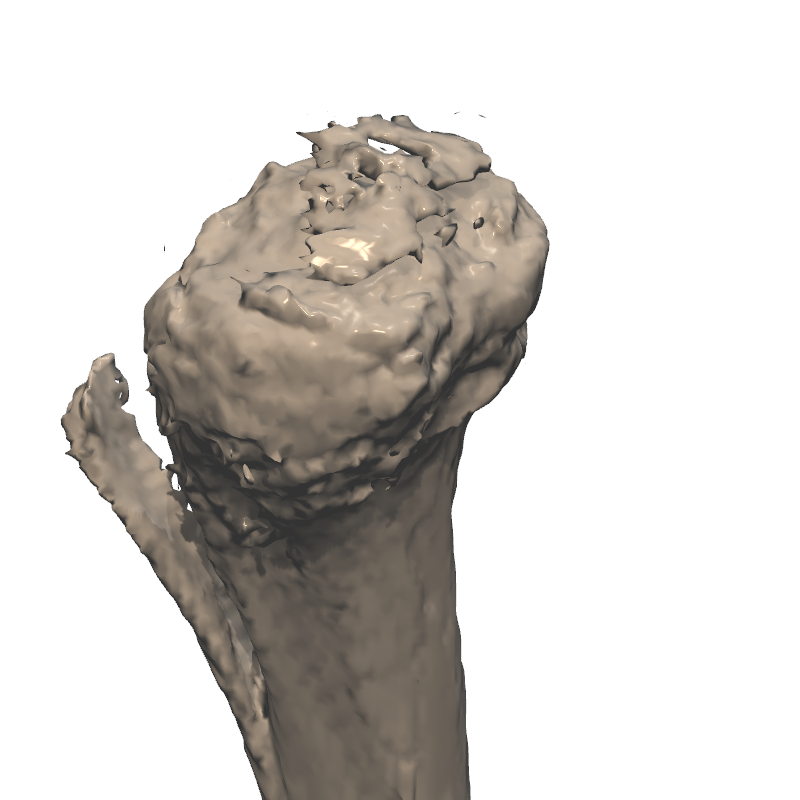} &
\includegraphics[width=0.10\linewidth,valign=c]{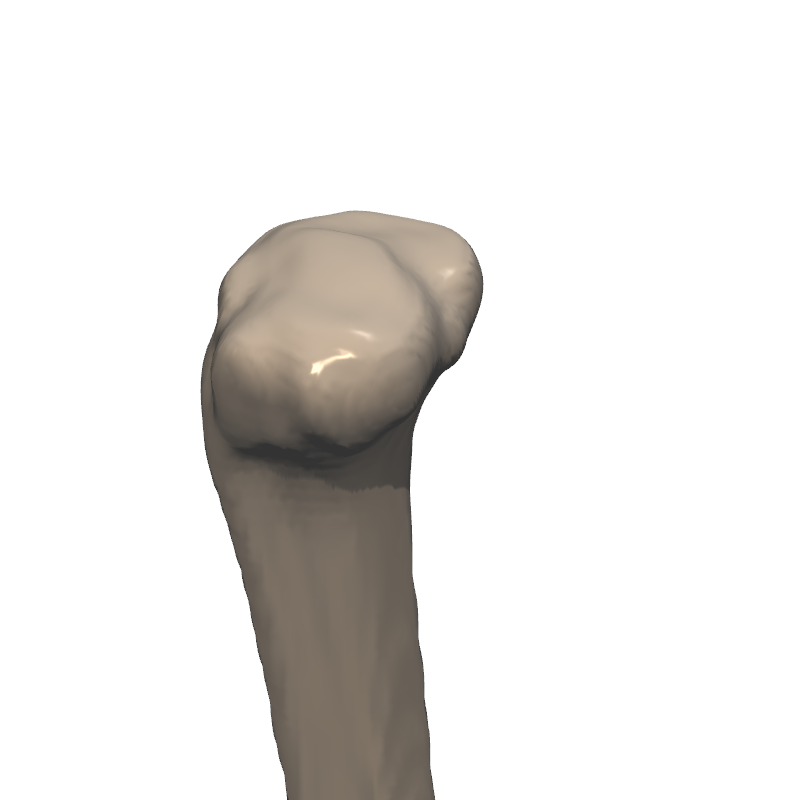} &
\includegraphics[width=0.10\linewidth,valign=c]{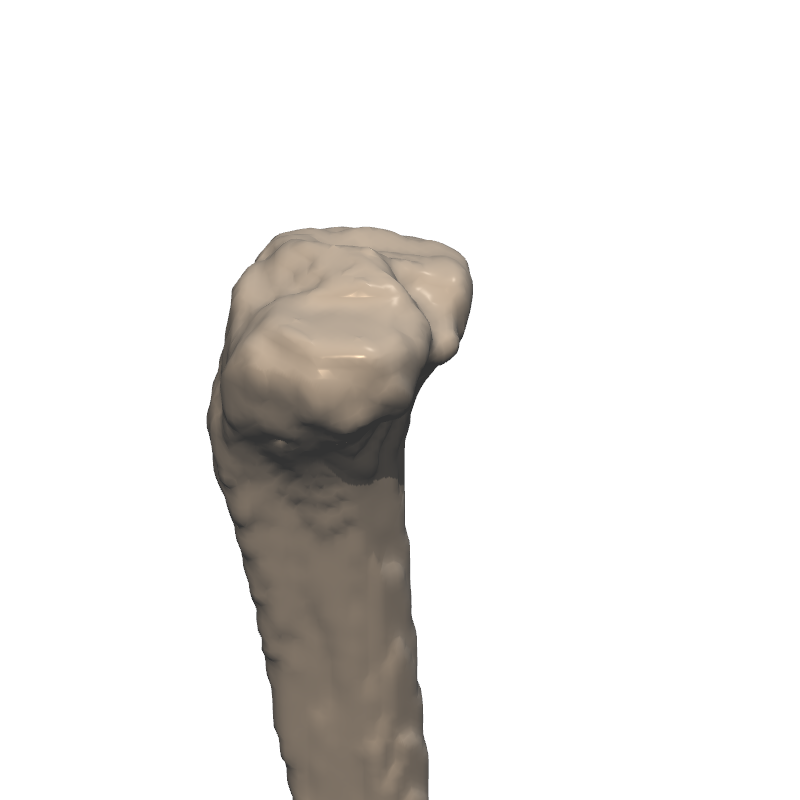} &
\includegraphics[width=0.10\linewidth,valign=c]{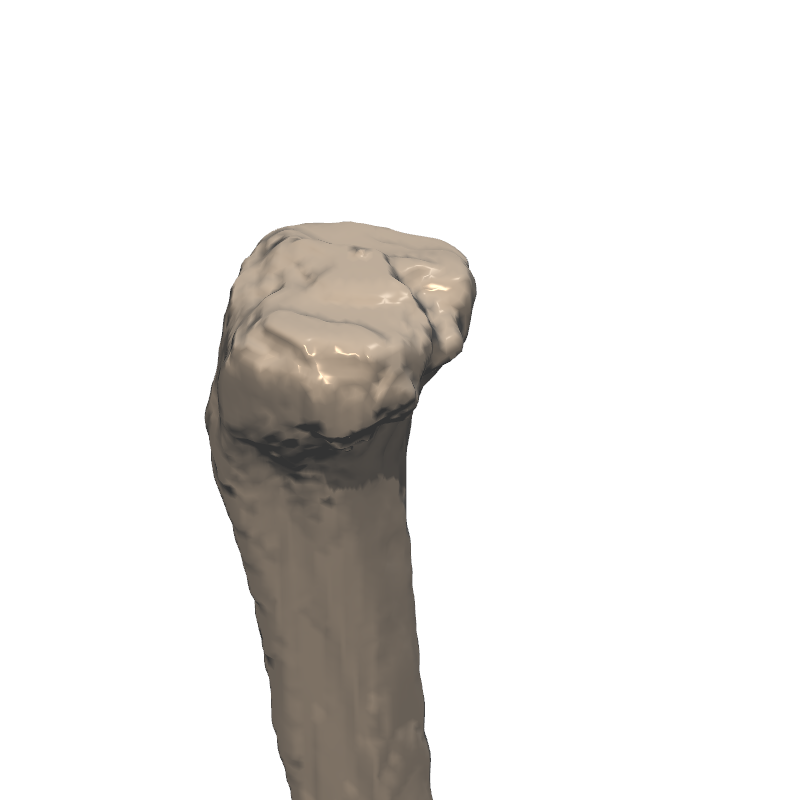} 
\\
\makecell{STN~\ref{eq_reg_p_locnet}\\Registration} & 
\includegraphics[width=0.10\linewidth,valign=c]{imgs/tibia_no_reg_gt.png} &
\includegraphics[width=0.10\linewidth,valign=c]{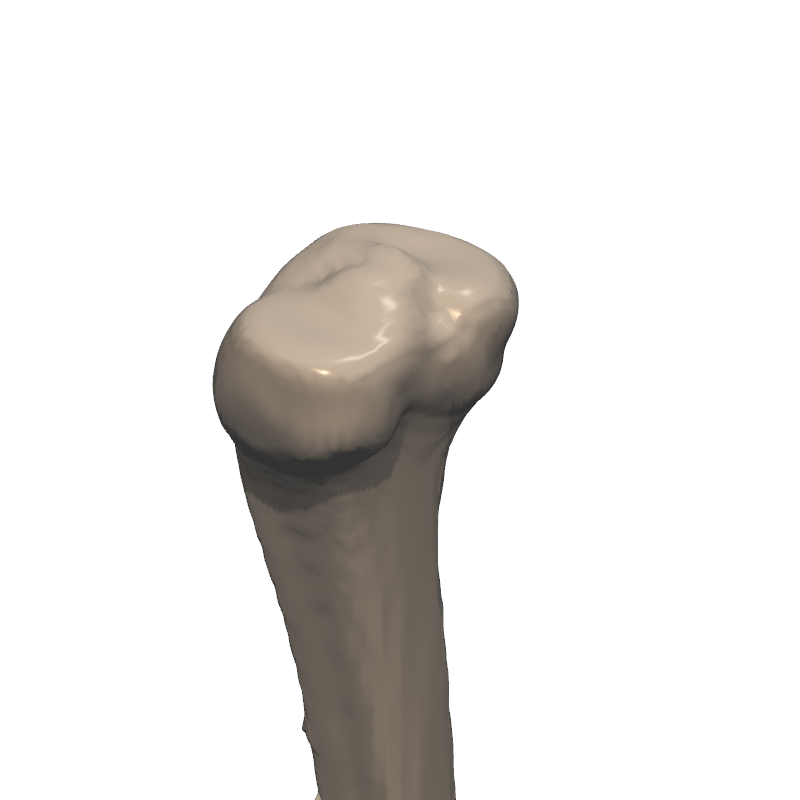} &
\includegraphics[width=0.10\linewidth,valign=c]{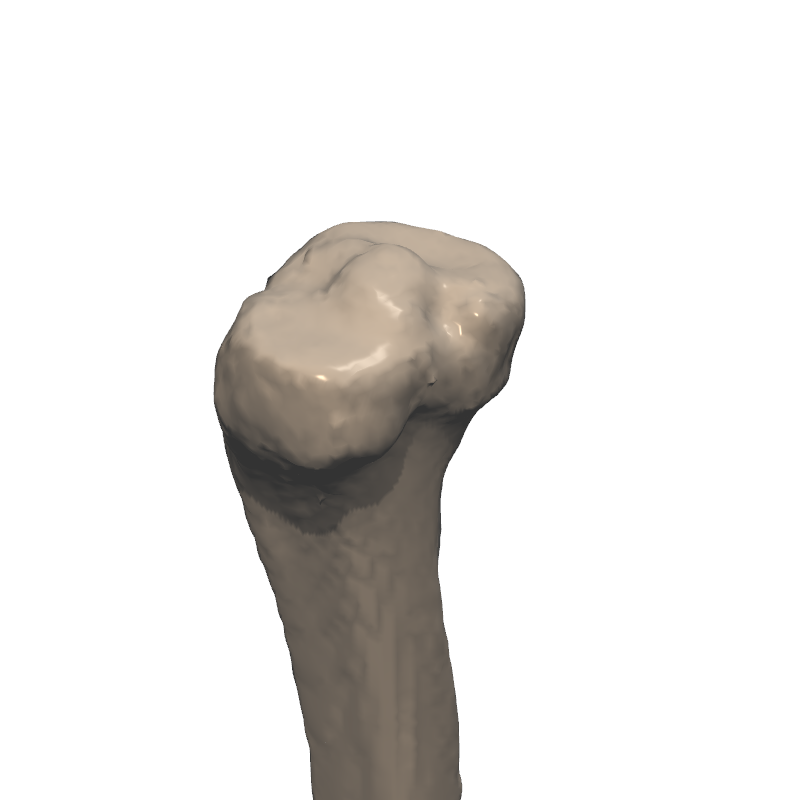} &
\includegraphics[width=0.10\linewidth,valign=c]{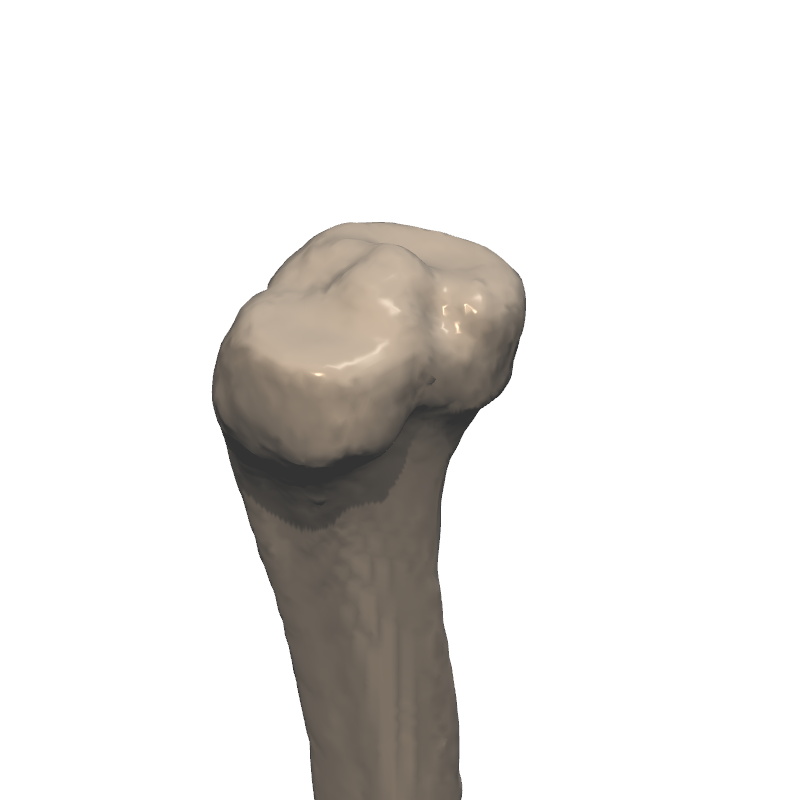} &
\includegraphics[width=0.10\linewidth,valign=c]{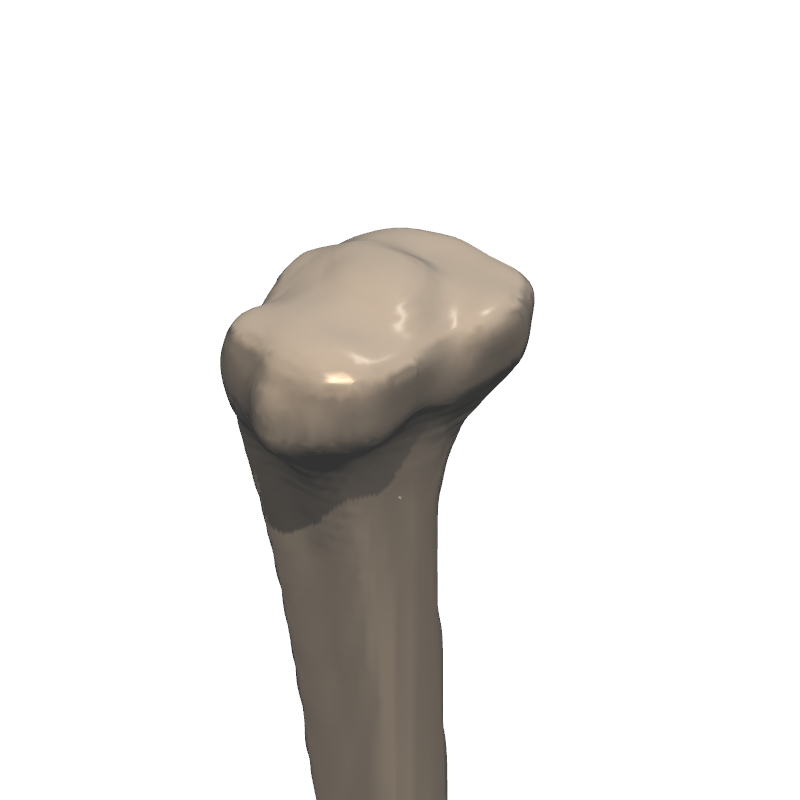} &
\includegraphics[width=0.10\linewidth,valign=c]{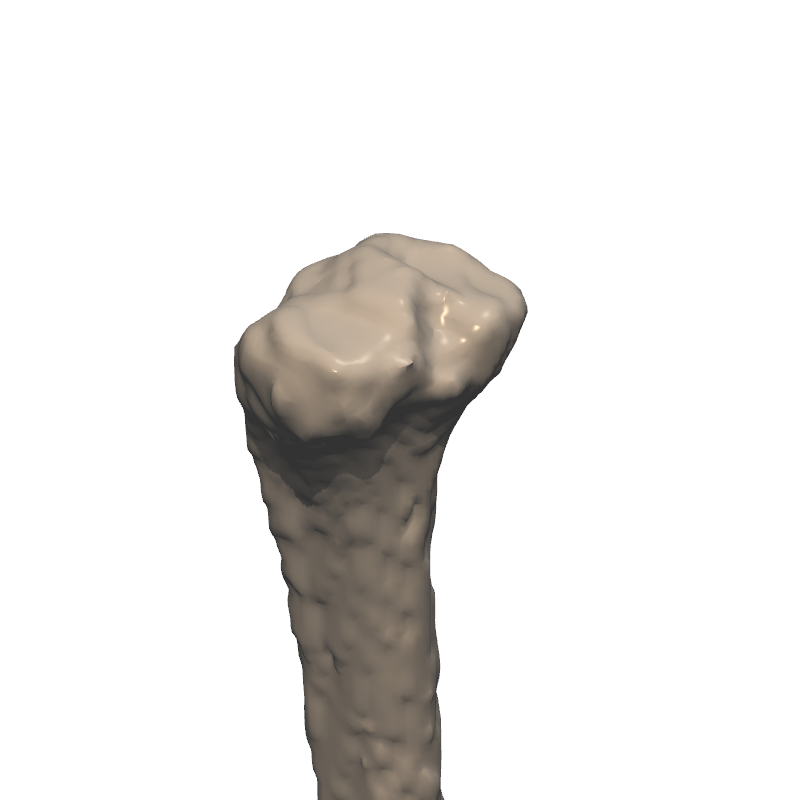} &
\includegraphics[width=0.10\linewidth,valign=c]{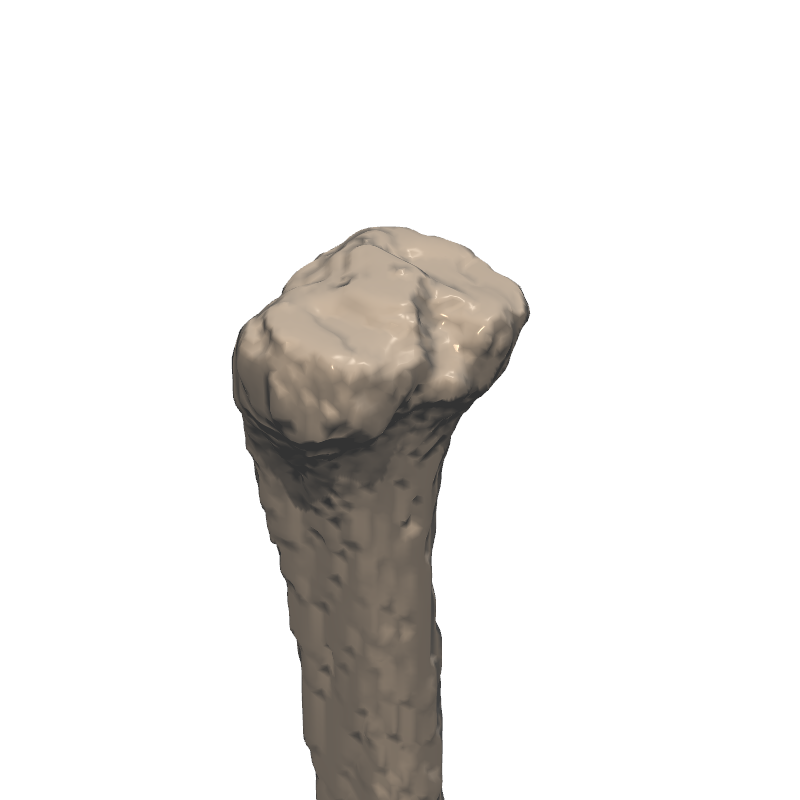}
\\
\hline
\hline
\makecell{Mask-robust\\STN~\ref{eq_frac_reg}\\Registration} & 
\includegraphics[width=0.10\linewidth,valign=c]{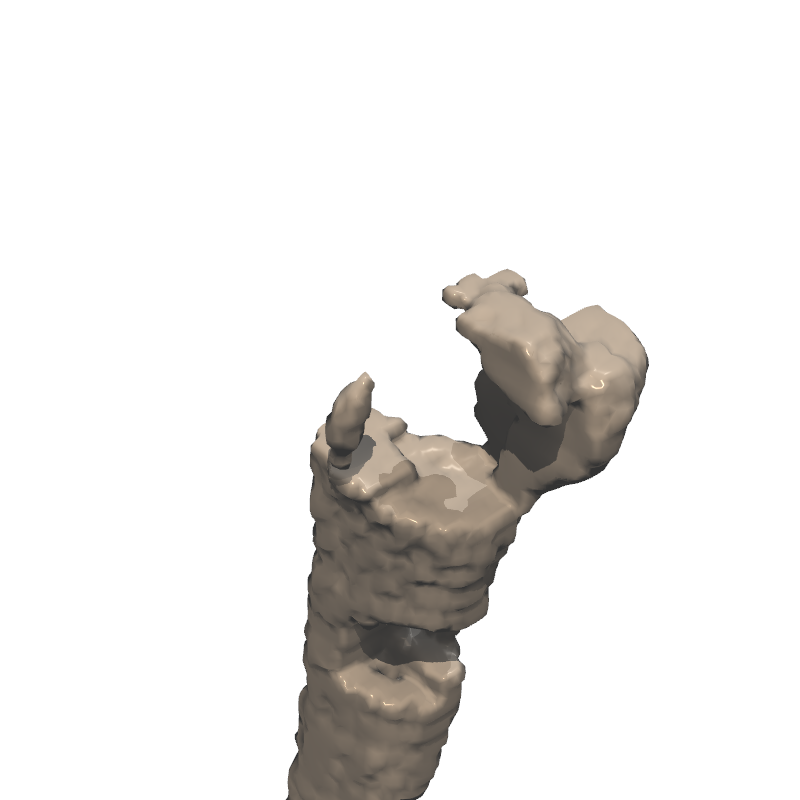} &   
N/A &
\includegraphics[width=0.10\linewidth,valign=c]{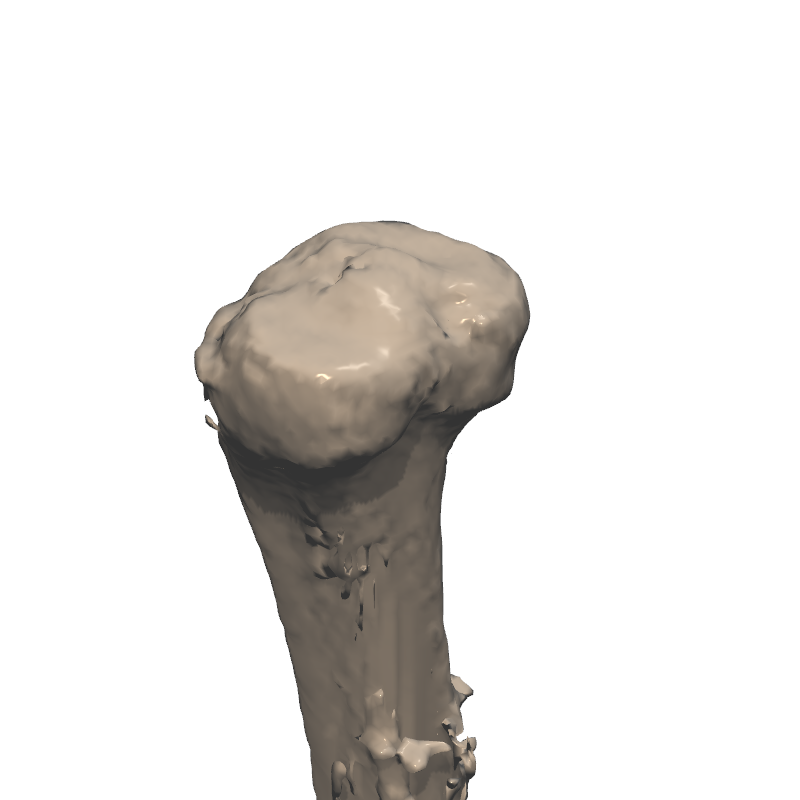} &
\includegraphics[width=0.10\linewidth,valign=c]{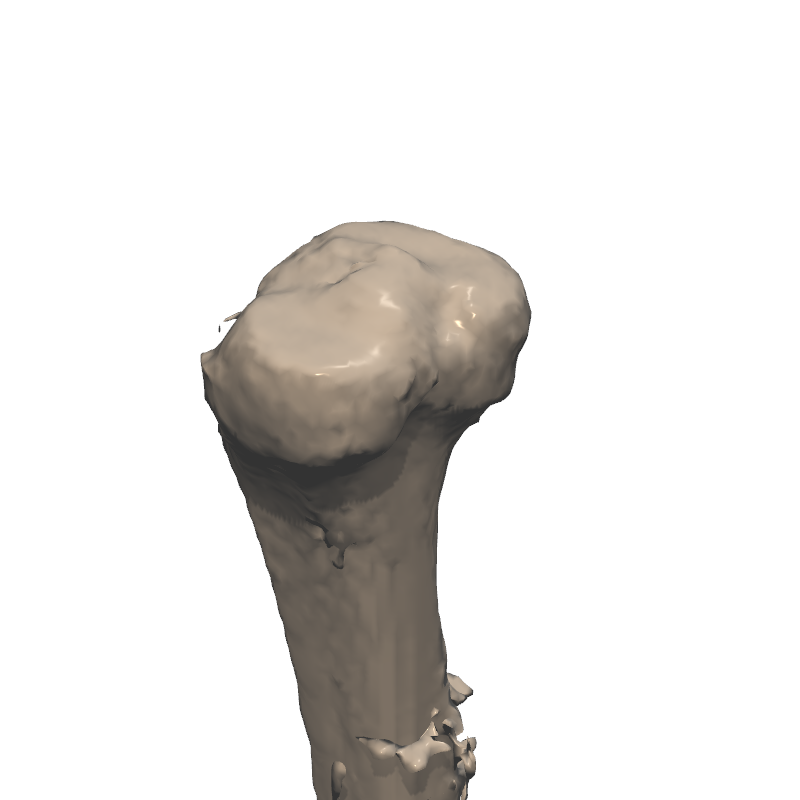} &
\includegraphics[width=0.10\linewidth,valign=c]{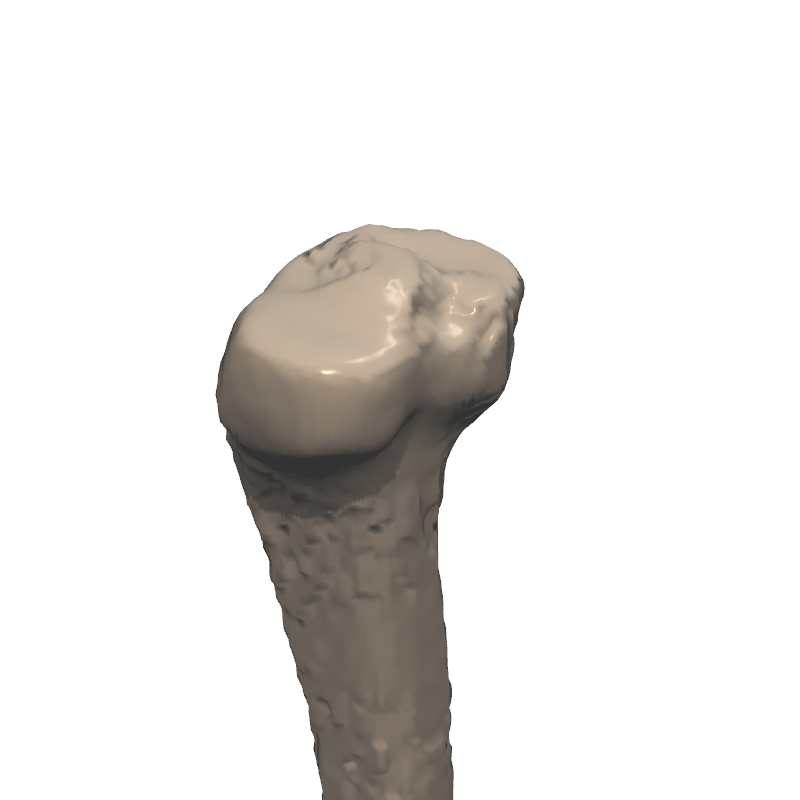} &
\includegraphics[width=0.10\linewidth,valign=c]{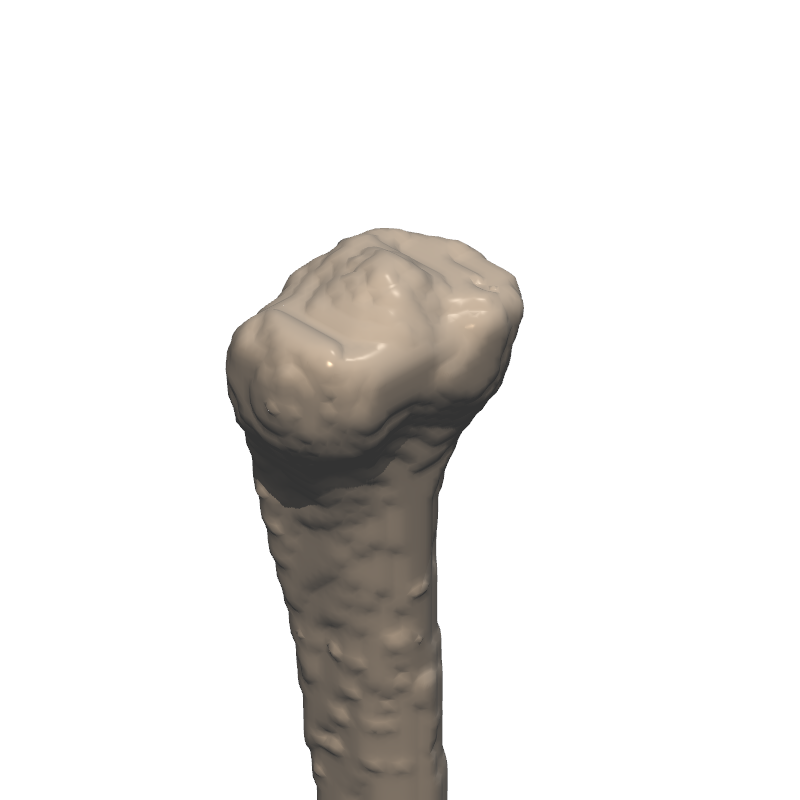} &
\includegraphics[width=0.10\linewidth,valign=c]{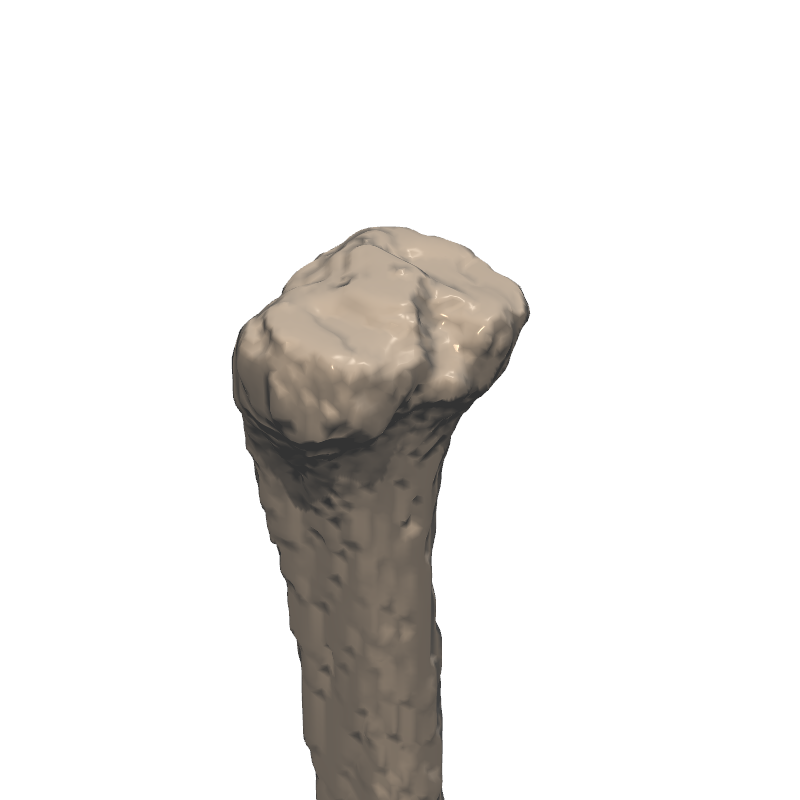}
\\[1.5ex]
\hline
\end{tabular}
\end{table}


\subsection{Data \& Training Details}
\label{sec:ex_pred_healthy}

\begin{anonymous}
    Our dataset consists of 249 lower-body CT scans\footnote{The study was approved by the institutional ethics board as part of the research project.}.
\end{anonymous}
\begin{disclosure}
    Our dataset consists of 249 full-body CT scans provided by the Department of Musculoskeletal Surgery at Charité\footnote{The study was approved by the institutional ethics board as part of the REPAIR research project. \url{https://repair.charite.de/}}.
\end{disclosure}
Using TotalSegmentator~\cite{wasserthal2023totalsegmentator} we get the data masks $\{m_i\}$ which crop out the tibia and its bounding box.
To make our training more robust, we re-sample these tibia CTs to a homogeneous voxel size of 1mm and finally use padding or cropping to make them isotropic, resulting in a tensor of the size of $x_i \in \mathbb{R}^{128 \times 128 \times 256}$ together with binary tibia voxel masks $m_i$ of the same dimension.

To increase the amount of available training data, we performed a
mirroring of the right-side skeleton along the median plane ($xy$ plane).  Including both legs, we
have in total 498 samples.  We used 442 samples as the training set,
and the remaining 56 samples as the validation set.

For the \emph{Direct Registration}~\eqref{eq_reg_p} we use the AdamW~\cite{loshchilov2017decoupled} optimizer with a learning rate of $1e^{-5}$, a batch size of 16, and let the optimizer run 50 steps for each batch and 100 epochs through the whole dataset. For the \emph{Registration Network}~\eqref{eq_reg_p_locnet} we use the AdamW optimizer with a learning rate of $1e^{-5}$, batch size of 16, and train the network for 1000 epochs.

For training the VAE we use the AdamW optimizer with a learning rate of $1e^{-5}$, the KL weight $\beta = 1e^{-2}$, and we train the network for 500 epochs for each data set. For VQ-VAE training we use the AdamW optimizer with a learning rate of $1e^{-5}$ and a quantizer weight $\beta$ following a step scheduler from $\beta_{\text{start}} = 0.5$ to $\beta_{\text{max}} = 1.0$ over the first 50 epochs, we train the network for 500 epochs for each data set. We use \emph{MSE} loss for all training task of AEs.

We train the mask-robust models using the same parameters but freeze the prototype parameters. 
For the training data, we use data that have been aligned using STN registration.

\subsection{Reconstruction Performance}

Table~\ref{tab:quat_comp_val_set} reports on quantitative reconstruction performances for the investigated registration and autoencoding methods on the validation data set. The \emph{w/o Registration} method naively applies autoencoders on the non-aligned raw data.
The \emph{Direct} and \emph{STN Registration} methods follow Sec.~\ref{sec:proto}.
The \emph{Mask-robust STN Registration} tests the mask-robust LocNet we trained for randomly masked and transformed data (Sec.~\ref{sec:data_aug}), but applied to the non-masked evaluation set as the other methods. The autoencoding method \emph{P} denotes the baseline of constantly predicting the prototype only (which is not available w/o registration).

As \textbf{metrics} we use the D-SSIM~\cite{wang2003multiscale} with varying window sizes $3, 7, 11, 15$,%
\footnote{The window size of D-SSIM  affects the scale of comparison, which smaller windows emphasizing finer-grained similarity, whereas larger windows focussing more on overall similarity. Evaluating multiple window sizes allows us to assess the reconstruction performance in terms of both detail and structure.} 
PSNR \cite{wang2009mean}, root mean squared error (RMSE) and root mean surface distance (RMSD). The latter denotes the distance between surfaces constructed using the marching cube algorithm~\cite{lorensen1998marching} with value window from $0.0$ to $1.0$ in millimeters.

Table~\ref{tab:quat_comp_val_set} shows results on the validation. 
Table~\ref{tab:qual_comp_val_set} more qualitatively illustrates
reconstructions for a sample CT for the different methods. The 3D
surface models were created using a multi-threshold marching cube
method.

\paragraph{Discussion:}

For the \emph{Direct AGS Registration}, the optimization process (\ref{eq_reg_p})
often converges to local optima, resulting in a prototype that fails
to converge to a unified skeletal model. 
In contrast, the trained \emph{STN} achieves a better solution, as indicated by the more uniform 3D reconstruction of the prototype.
This superiority is further supported by the quantitative analysis on
both the training and validation datasets, where the registration
results obtained using the STN (\ref{eq_reg_p_locnet}) consistently outperform the
direct parameter optimization method (\ref{eq_reg_p}).



When comparing autoencoder methods, the PCA results clearly indicate that registration significantly enhances reconstruction quality.
In constrast, the VAE and VQ-VAE methods are more robust to inaccurate registration.
The quantitative results show that the combination of using the trained STN registration with a (VQ-)VAE encoder consistently outperforms the alternatives.

The lower blocks of tables \ref{tab:quat_comp_val_set} and \ref{tab:qual_comp_val_set} show results when input CTs were randomly masked -- only the trained mask-robust STN registration is applicable. 
Our results indicate that reconstruction results are still of high quality despite input masking.
Overall we find:
\begin{itemize}
    \item Without masking: VQ-VAE achieves better quantitative performance on unmasked data, as the use of discrete latent embeddings allows it to memorize high-frequency structural patterns. These hard assignments enable the decoder to reconstruct sharper edges and localized anatomical details more accurately compared to the continuous latent codes of a VAE.
    \item With masking: In contrast, the continuous latent space of a VAE facilitates better generalization to unseen or partially missing inputs. During inpainting, the smoothness of the latent manifold allows the decoder to interpolate plausible structures, leading to more anatomically coherent results and fewer artifacts, especially under large fracture masks.
    \item Artifacts in VQ-VAE with masking arise due to limited codebook flexibility, making it struggle with unseen or missing patterns. We hypothesize that increasing the codebook size and applying EMA updates can reduce hallucinations by improving coverage of anatomical variations. Additionally, encouraging balanced codeword usage through entropy regularization may lead to more reliable reconstructions under masked inputs.
\end{itemize}

\begin{figure}[t]
    \centering
    \begin{tabularx}{\linewidth}{c@{}c @{\hspace{2\tabcolsep}} c@{}c @{\hspace{2\tabcolsep}} c@{}c}
        \includegraphics[width=0.15\linewidth,valign=c]{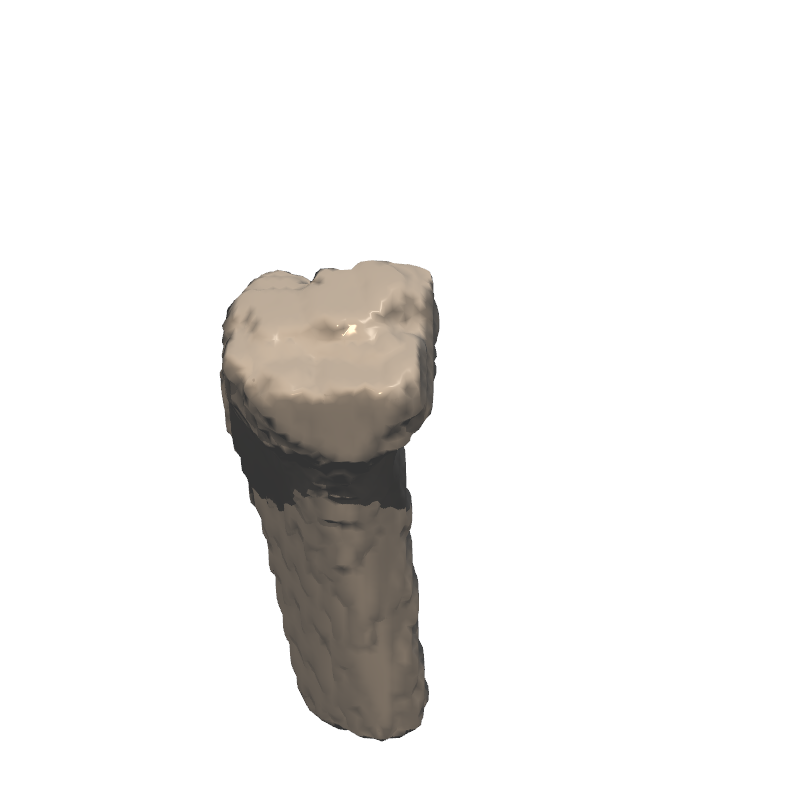} &
        \includegraphics[width=0.15\linewidth,valign=c]{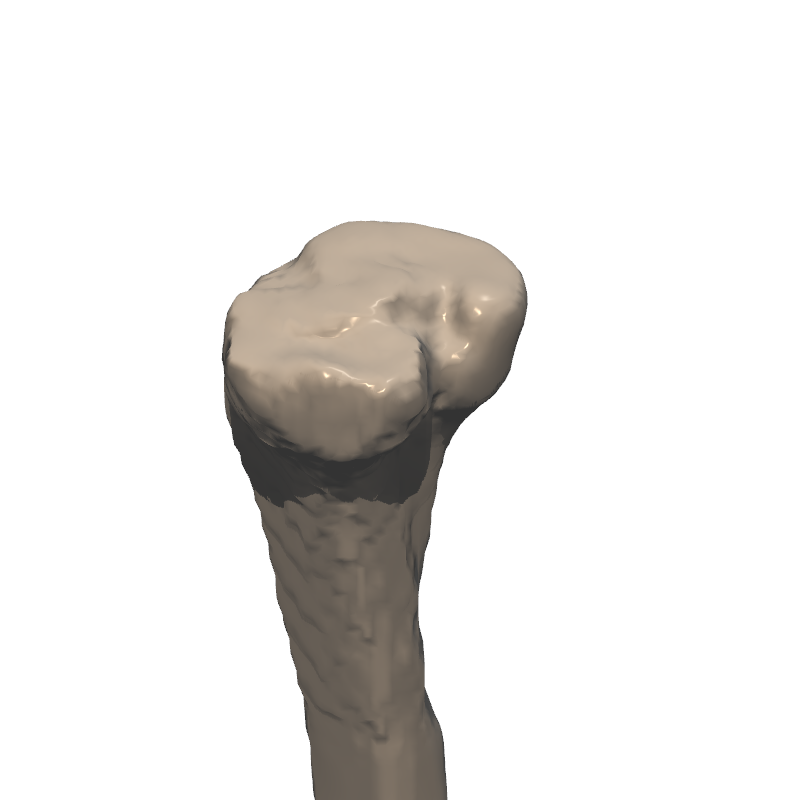} &
        \includegraphics[width=0.15\linewidth,valign=c]{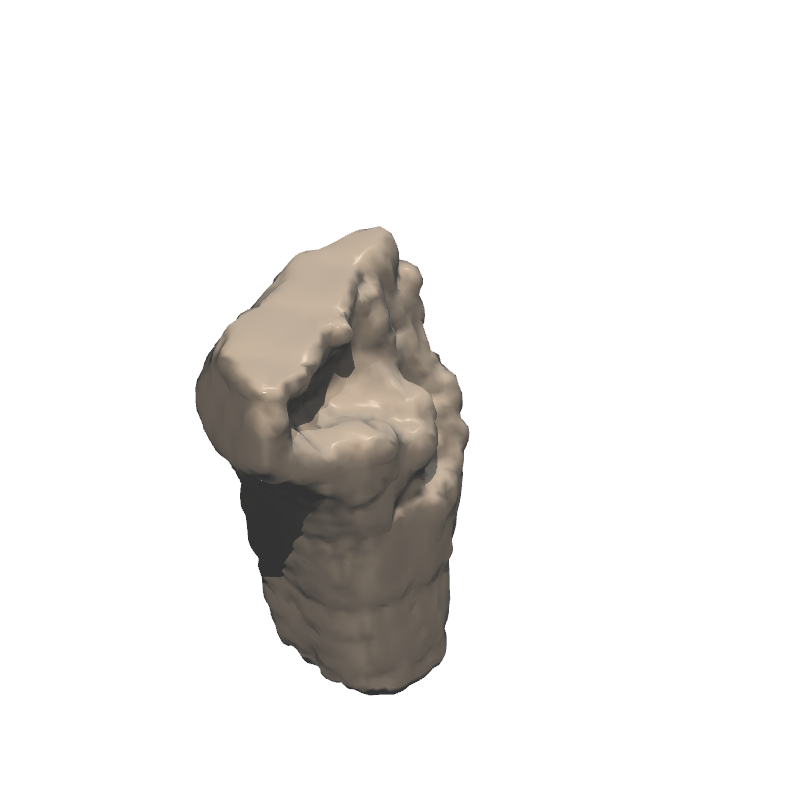} &
        \includegraphics[width=0.15\linewidth,valign=c]{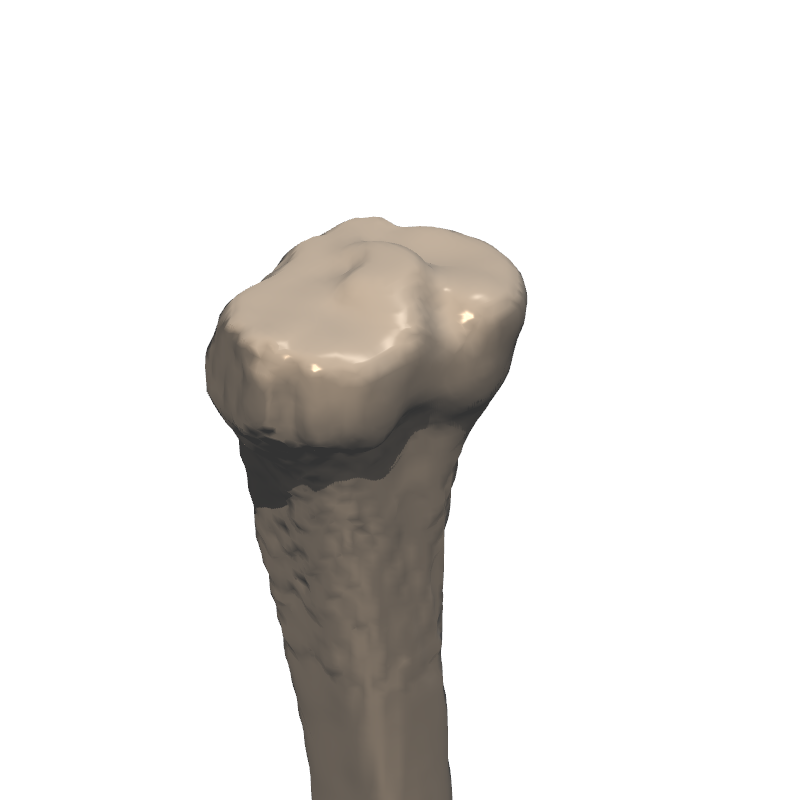} &
        \includegraphics[width=0.15\linewidth,valign=c]{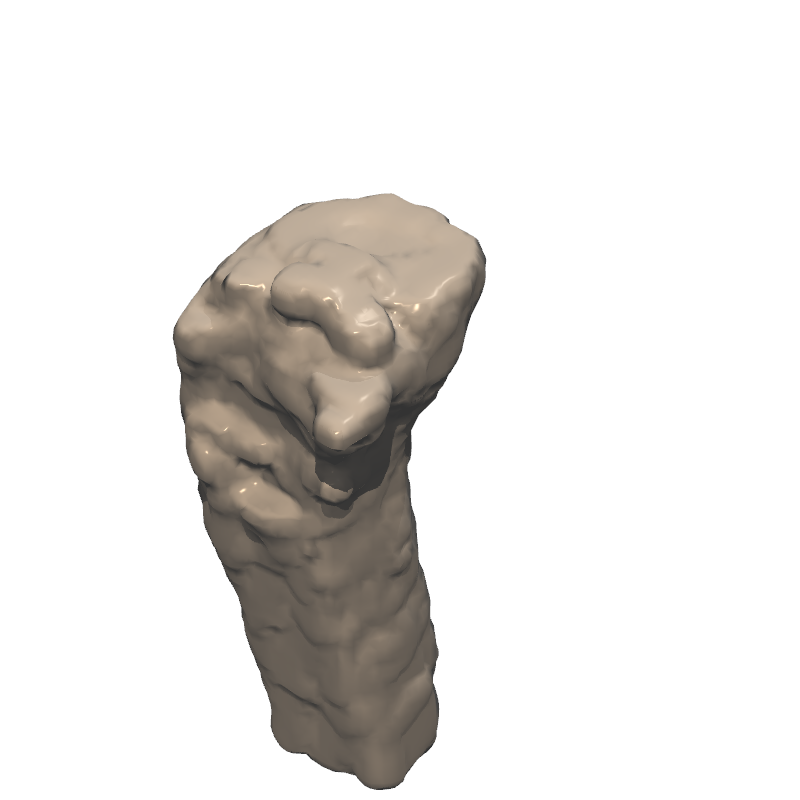} &
        \includegraphics[width=0.15\linewidth,valign=c]{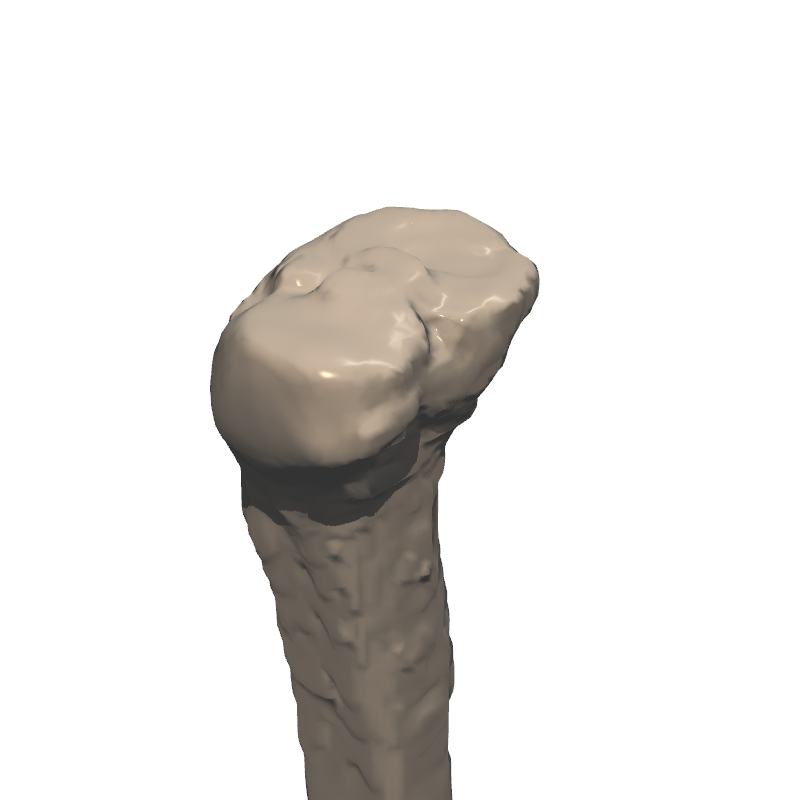} \\
        Input & Reconstruction & Input & Reconstruction & Input & Reconstruction \\
    \end{tabularx}
    \caption{Application of our mask-robust STN+VAE pipeline on three exemplary CTs where fractures have been masked out.}
    \label{fig:qualitative_comp}
\end{figure}

Finally, Fig.~\ref{fig:qualitative_comp} presents results on applying our reconstruction pipeline on three exemplary fracture-masked CTs.


\section{Conclusion}
Our work demonstrates that robust predictive reconstruction of a healthy tibia from fractured CT images is achievable by combining deep models for spatial registration and autoencoders. 
Specifically, by introducing a modified STN for consistent registration, evaluating alternative autoencoder models, and introducing a novel pipeline to train mask-robust models we achieve high-quality reconstructions also for fracture-masked CTs. 
Our experimental results indicate that while VQ-VAE delivers sharper details in unmasked scenarios, the continuous latent space of the VAE offers enhanced performance when handling masked inputs, effectively reducing reconstruction artifacts. 
Our combination of STN with VAEs not only improves the fidelity of patient-specific skeletal models but also sets the stage for integrating such techniques into surgical planning workflows. 

\bibliographystyle{splncs04}
\bibliography{citations}


\end{document}